\documentclass[conference,a4paper,10pt]{IEEEtran}\usepackage[]{graphicx}\usepackage[]{color}
\makeatletter
\def\maxwidth{ %
  \ifdim\Gin@nat@width>\linewidth
    \linewidth
  \else
    \Gin@nat@width
  \fi
}
\makeatother

\definecolor{fgcolor}{rgb}{0.345, 0.345, 0.345}

\usepackage{framed}
\makeatletter
 {\par\unskip\endMakeFramed%
 \at@end@of@kframe}
\makeatother

\definecolor{shadecolor}{rgb}{.97, .97, .97}
\definecolor{messagecolor}{rgb}{0, 0, 0}
\definecolor{warningcolor}{rgb}{1, 0, 1}
\definecolor{errorcolor}{rgb}{1, 0, 0}
\newenvironment{knitrout}{}{} % an empty environment to be redefined in TeX

\usepackage{alltt}

\usepackage{amsmath}
\usepackage{amsfonts}
\newcommand{\Xe}{\mathbf{X}_\textrm{e}}
\IfFileExists{upquote.sty}{\usepackage{upquote}}{}
\begin{document}

\title{Optimistic Semi-supervised \\ Least Squares Classification}

\author{\IEEEauthorblockN{Jesse H. Krijthe\IEEEauthorrefmark{1}\IEEEauthorrefmark{3},
Marco Loog\IEEEauthorrefmark{1}\IEEEauthorrefmark{2}}
\IEEEauthorblockA{jkrijthe@gmail.com, m.loog@tudelft.nl\\ \IEEEauthorrefmark{1}Pattern Recognition Laboratory, Delft University of Technology, The
Netherlands}
\IEEEauthorblockA{\IEEEauthorrefmark{2}The Image Group, Department of Computer Science, University of Copenhagen, Denmark}
\IEEEauthorblockA{\IEEEauthorrefmark{3}Department of Molecular Epidemiology, Leiden University Medical Center, The Netherlands}}

\maketitle

\begin{abstract}
The goal of semi-supervised learning is to improve supervised classifiers by using additional unlabeled training examples. In this work we study a simple self-learning approach to semi-supervised learning applied to the least squares classifier. We show that a soft-label and a hard-label variant of self-learning can be derived by applying block coordinate descent to two related but slightly different objective functions. The resulting soft-label approach is related to an idea about dealing with missing data that dates back to the 1930s. We show that the soft-label variant typically outperforms the hard-label variant on benchmark datasets and partially explain this behaviour by studying the relative difficulty of finding good local minima for the corresponding objective functions.
\end{abstract}

\IEEEpeerreviewmaketitle

\section{Introduction}
Semi-supervised learning aims to improve supervised classifiers by incorporating abundant unlabeled training examples in the training of a classifier. Many approaches to semi-supervised learning have been suggested (for one overview see \cite{Chapelle2006}), mostly taking advantage of assumptions that allow information about the distribution of the feature values to inform decisions as to what constitutes a good classifier.

Instead of introducing yet another assumption to the literature, the goal of this work is to revisit a classic approach to semi-supervised learning to incorporate unlabeled data in the training of the least squares classifier: self-learning. While self-learning is often proposed as an ad hoc procedure, we properly formulate it as the minimizer of a particular objective function in the least squares setting. A slightly different formulation then leads to a new type of soft-label self-learning. 

This soft-label self-learning procedure is closely related to a procedure that dates all the way back to an approach to missing data in least squares regression by Yates \cite{Yates1933} and Healy and Westmacott \cite{Healy1956} from the 1930s and 1950s respectively. We revisit these ideas in the context of the present paper.

In a set of experiments, we show that the soft-label self-learning variant tends to outperform the hard-label variant. We explain this behaviour based on the differing difficulty of finding good local minima of the corresponding objective functions. 

The paper is structured as follows: we will first explain how our ideas relate to early approaches to deal with missing data in statistics and to the well known self-learning and expectation maximization (EM) approaches to semi-supervised learning. Afterwards we will formulate two different objective functions for the semi-supervised least squares classification problem. We show that applying block coordinate descent \cite[Ch. 2.7]{Bertsekas1999} to these objective functions corresponds to either hard-label, respectively soft-label self-learning. We then study the properties of these objective functions using some simulation studies and end with a set of experiments on benchmark datasets that shows that soft-label self-learning tends to outperform the hard-label variant.

\section{Historical Perspective}

\subsection{Self-learning and EM}
Arguably, the most simple and straightforward approach to semi-supervised learning is a procedure known as self-learning. Starting with the supervised classifier trained using only labeled data, one predicts the labels for the unlabeled objects and uses these in the next iteration of training the classifier. This is done until the predicted labels of the unlabeled objects no longer change. Self-learning was introduced in \cite{McLachlan1975} and \cite{McLachlan1977} as a more feasible alternative to the proposal by \cite{Hartley1968b} to consider all possible labelings of the unlabeled data and find the one that minimizes the log likelihood of an ANOVA model.

For probabilistic models, such as linear discriminant analysis based on a Gaussian model of each class, a closely related approach to self-learning is to apply the Expectation Maximization (EM) procedure \cite{Dempster1977} that attempts to find the parameters that maximize the likelihood  of the model after marginalizing out the unknown labels. This leads to an algorithm where in each iteration, the parameters of the model are estimated based on the current partial assignment of objects to different classes (the maximization step), after which these assignments are updated based on the new model parameters (the expectation step). The assignments of unlabeled objects to classes in each stage is a partial assignment based on the posterior probability given by the current parameter estimates of the model, rather than a hard assignment to a single class as is typically used in self-learning.

\subsection{Connection to Missing Data in Regression}
In this work we consider the least squares classifier, the classifier that minimizes the squared error of the fit of a (linear) decision function to the class labels encoded as $\{0,1\}$ \cite[p.103]{Hastie2009}. While the classifier in this form does not have a proper likelihood function to which we can apply the EM procedure, it is closely related to least squares regression where Healy and Westmacott \cite{Healy1956} proposed a similar iterative procedure. The idea behind this iterative procedure dates back as early as 1933 to work by Yates \cite{Yates1933}. Yates noted that one can get unbiased estimates of the regression parameters when outcomes are missing by plugging in the regression estimates for the missing values that were given by the complete data alone. The problem is how to find these regression estimates. While for some regression designs analytical solutions exist \cite{Wilkinson1958} to fill in the missing values, \cite{Healy1956} describes the simple approach of starting with random values and iteratively updating the regression parameters and setting the missing values to the values predicted by the updated model.
They prove this procedure reduces the residual sum of squares at each iteration. It was later shown \cite{Dempster1977} that for the linear regression model with Gaussian errors, this procedure corresponds to the EM algorithm.

Note that the goal of these procedures is to deal with missing values in a convenient automatic way, not necessarily to improve the estimator \cite[Ch. 2]{Little2002}, as is the case in semi-supervised learning.

% Preece builds upon this work by suggesting larger updates of the missing values in each iteration to speed up convergence. This is based on the observation that the procedure proposed by Yates for a single missing value uses a larger update than the simple scheme by Healy and Westmacott. 

Unlike in the regression setting, in least squares classification, we know that the true labels are not values in $\mathbb{R}$ but rather restricted to $\{0,1\}$, or as we will consider, $[0,1]$. We will show that the iterative approach of \cite{Healy1956}, when taking into account the constraints that missing outcomes are $[0,1]$ can be formulated as a block coordinate descent procedure on the quadratic loss function and that, in fact, an improved classifier can generally be obtained by using this procedure. For a different take on how to introduce soft-labels in self-learning and how this relates to EM, see \cite{Mey2016}.

\subsection{Pessimism vs. Optimism}
To guard against unlabeled data leading to worse classifiers one can attempt to construct a classifier that improves over the supervised classifier even for the worst possible labeling of the unlabeled objects, which is proposed in contrastive pessimistic learning, see \cite{Loog2016} and \cite{Krijthe2015}. This is the pessimistic approach to semi-supervised learning. In contrast, we refer to the approaches studied here as optimistic approaches, where in each step of the algorithm, we consider the best-case labels, rather than the worst-case labels.

\section{Regularized Least Squares Classification}
Let $\mathbf{x}$ denote a $d\times 1$ feature vector, potentially containing a constant feature encoding the intercept.
$\mathbf{X}$ is the $L \times d$ design matrix of the labeled examples, while $\mathbf{X}_\text{u}$ is the  $U \times d$ design matrix of the unlabeled examples. $\mathbf{y}$ is the $L\times 1$ vector containing the labels encoded as $\{0,1\}$.
$\mathbf{w}$ denotes the weight vector of a linear classifier.

In the supervised setting, the binary regularized least squares classifier is defined by the following loss function:
\begin{equation}
J_s(\mathbf{w}) = \sum_{i=1}^L (\mathbf{x}_i^\top \mathbf{w}-y_i)^2 + \lambda \|\mathbf{w} \|^2 \,. \label{eq:supervisedobjective}
\end{equation}
The minimizer of this objective function has a well-known closed form solution:
\begin{equation}
\mathbf{w} = \left( \mathbf{X}^\top \mathbf{X}  + \lambda \mathbf{I} \right)^{-1} \mathbf{X}^\top \mathbf{y} \,. \label{eq:olssolution}
\end{equation}
To label a new object a threshold is typically set at $\tfrac{1}{2}$:
$$
c_\mathbf{w}(\mathbf{x}) = 
\begin{cases}
    1, & \mathbf{x}^\top \mathbf{w}>\tfrac{1}{2}\\
    0,              & \text{otherwise}
\end{cases}
$$

\section{Optimistic Semi-supervised LSC}

\begin{knitrout}
\definecolor{shadecolor}{rgb}{0.969, 0.969, 0.969}\color{fgcolor}\begin{figure}
\includegraphics[width=\maxwidth]{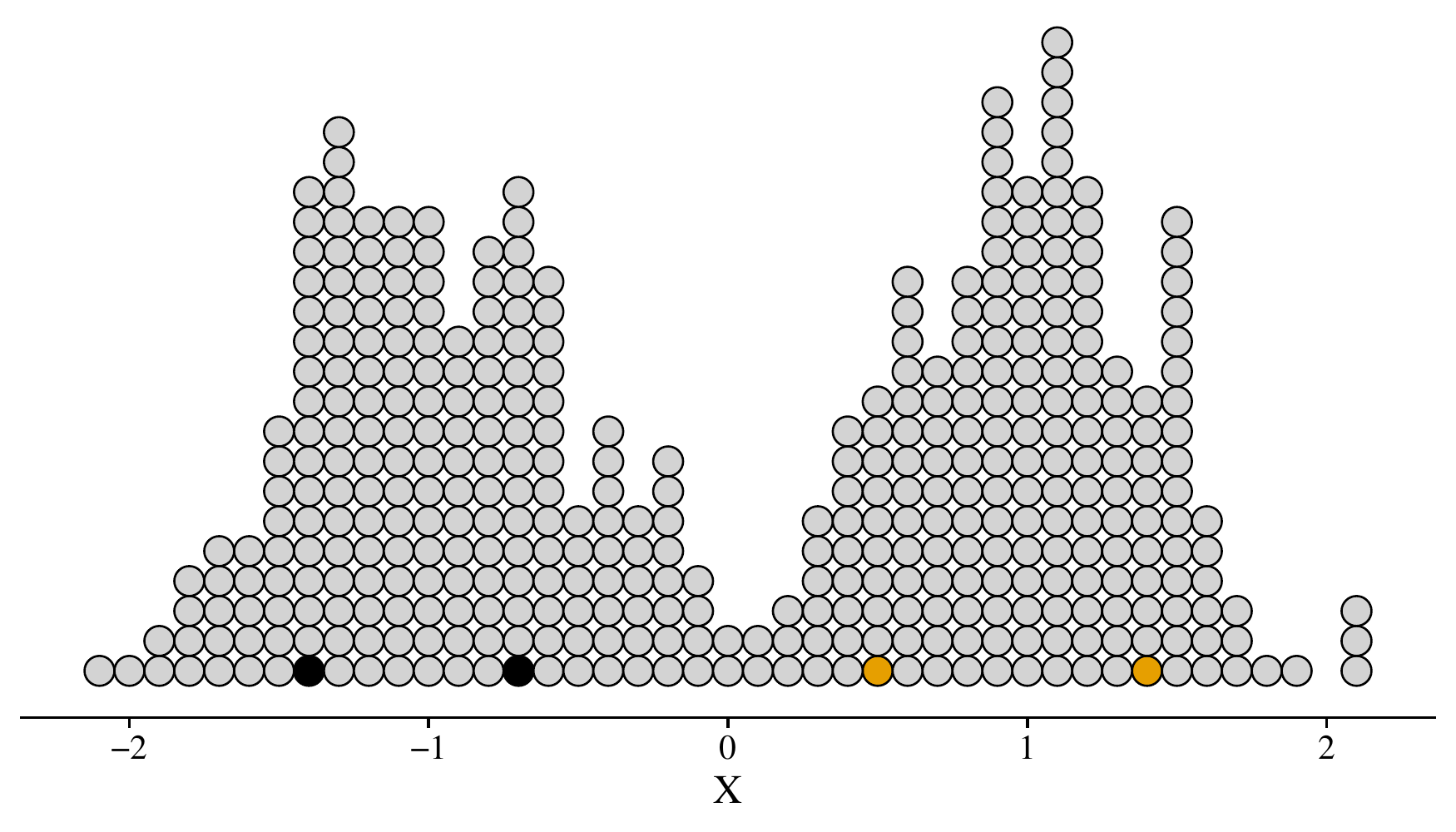} \caption[Dotplot of example dataset with 4 labeled objects and 396 unlabeled objects]{Dotplot of example dataset with 4 labeled objects and 396 unlabeled objects.}\label{fig:attractionexample}
\end{figure}

\end{knitrout}

We present two closely related ways to extend the supervised least squares objective function to include unlabeled data in an optimistic way. We refer to the first formulation as the label based formulation and show that applying a block coordinate descent procedure to this objective function corresponds to a type of soft-label self-learning, similar to Expectation Maximization. The second formulation is a very similar responsibility based formulation where a block coordinate descent approach corresponds to the familiar hard-label self-learning.

\subsection{Label Based Formulation}
A straightforward way to include the unlabeled objects into the objective function \eqref{eq:supervisedobjective} is to introduce a variable for the missing labels $\mathbf{u}$ and then use the supervised objective function as if we knew these labels:
$$
J_l(\mathbf{w},\mathbf{u}) = \| \Xe \mathbf{w}-\begin{bmatrix} \mathbf{y} \\ \mathbf{u} \end{bmatrix} \|^2 + \lambda \|\mathbf{w} \|^2
$$
where $\Xe$ is the concatenation of $\mathbf{X}$ and $\mathbf{X}_\text{u}$. Since we do not know $\mathbf{u}$, one possibility is to minimize over both $\mathbf{w}$ and $\mathbf{u}$:
$$
\min_{\mathbf{w}\in \mathbb{R}^d,\mathbf{\mathbf{u} \in [0,1]^U}} J_l(\mathbf{w},\mathbf{u}) \,.
$$
Taking the gradient with respect to $\mathbf{w}$ and setting it equal to zero gives
$$
\mathbf{w} = \left( \mathbf{X}_\text{e}^\top \mathbf{X}_\text{e}  + \lambda \mathbf{I} \right)^{-1} \mathbf{X}_\text{e}^\top \mathbf{y}_\text{e} \,.
$$
% $$
% \frac{\nabla J_l(\mathbf{w},\mathbf{u})}{\nabla \mathbf{w}} = 2 \mathbf{X}_\text{e}^\top \mathbf{X}_\text{e} \mathbf{w} - 2 \mathbf{X}_\text{e}^\top \mathbf{y}_\text{e} + 2 \lambda \mathbf{w}
% $$
% $$
% \frac{\nabla J_l(\mathbf{w},\mathbf{u})}{\nabla \mathbf{w}} = 2 \mathbf{X}_\text{e}^\top \mathbf{X}_\text{e} \mathbf{w} - 2 \mathbf{X}_\text{e}^\top \mathbf{y}_\text{e} + 2 \lambda \mathbf{w}
% $$
where $\mathbf{y}_\text{e}=\begin{bmatrix} \mathbf{y} \\ \mathbf{u} \end{bmatrix}$. So given a choice of labels $\mathbf{u}$ for the unlabeled objects, this naturally has the same form as the solution \eqref{eq:olssolution} in the supervised setting.

The gradient with respect to the unknown labels is:
$$
\frac{\nabla J_l(\mathbf{w},\mathbf{u})}{\nabla \mathbf{u}} = -2 (\mathbf{X}_\text{u} \mathbf{w} - \mathbf{u})
$$
so given a set of weights, $\mathbf{w}$, the minimizing labeling is 
$$\mathbf{u} = \mathbf{X}_\text{u} \mathbf{w} \,.$$
Taking into account, however, the constraint that $\mathbf{u} \in [0,1]^U$, the solution is to project each label onto $[0,1]$:
$$
u_i = 
\begin{cases}
0, & \text{if } \mathbf{x}_i^\top \mathbf{w} < 0\\
\mathbf{x}_i^\top \mathbf{w}, & \text{if } 0 < \mathbf{x}_i^\top \mathbf{w} < 1\\
1, & \text{if } \mathbf{x}_i^\top \mathbf{w} > 1
\end{cases}
$$
In this formulation, for a given $\mathbf{w}$ we get hard labelings whenever the decision function gives a value outside $(0,1)$, but a soft assignment, between $(0,1)$, if it does not.

\begin{knitrout}
\definecolor{shadecolor}{rgb}{0.969, 0.969, 0.969}\color{fgcolor}\begin{figure*}
\includegraphics[width=\maxwidth]{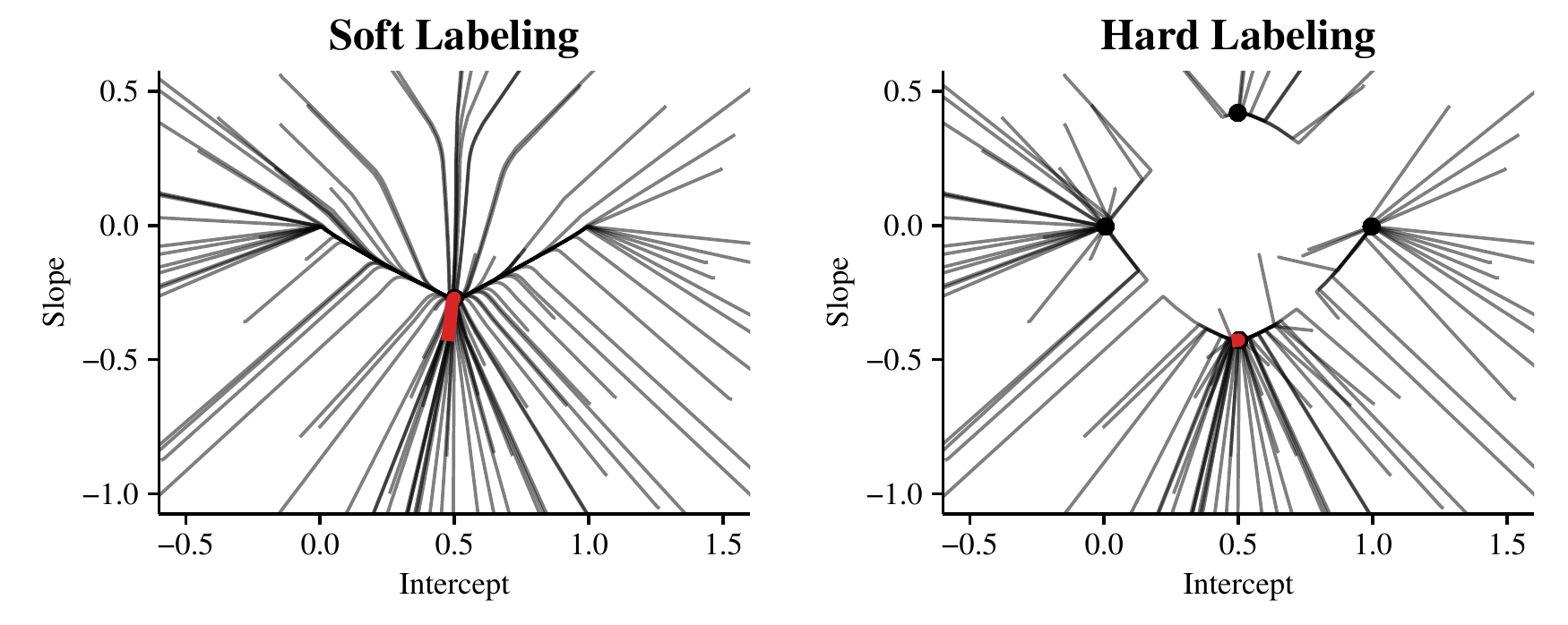} \caption[Convergence paths of the block coordinate descent procedure for different starting values for the label based objective function (left) and responsibility based objective function (right)]{Convergence paths of the block coordinate descent procedure for different starting values for the label based objective function (left) and responsibility based objective function (right). Shown are the weights corresponding to the intercept and single feature of a classifier applied to the problem shown in Figure 1. Shown in red is the optimization path starting with the supervised solution.}\label{fig:attraction1d}
\end{figure*}

\end{knitrout}

\subsection{Responsibility Based Formulation}
A different way to include the unlabeled data is by introducing a variable $\mathbf{q} \in [0,1]^U$ that indicates what portion of the loss of the two classes should be attributed to the loss for each unlabeled object. We refer to this $\mathbf{q}$ as a vector of responsibilities. While the responsibility is closely related to soft-labels in the previous section, it leads to a slightly different loss function. For generality, let $m$ be the numerical encoding used for one class and $n$ for the other. In this work, we shall use $m=1$ and $n=0$. The objective function becomes:
\begin{align}
J_r(\mathbf{w},\mathbf{q}) = & \| \mathbf{X} \mathbf{w}-\mathbf{y} \|^2 + \lambda \|\mathbf{w} \|^2 \nonumber \\
& + \sum_{i=1}^{U}  q_i (\mathbf{x}_i^\top \mathbf{w} - m)^2  + (1-q_i) (\mathbf{x}_i^\top \mathbf{w} - n)^2 \,. \nonumber \\ \nonumber
\end{align}
Now to solve $\min_{\mathbf{w} \in \mathbb{R}^d, \mathbf{\mathbf{q} \in [0,1]^U}} J_r(\mathbf{w},\mathbf{q})$ we again consider the gradients
$$
\frac{\nabla J_r}{\nabla \mathbf{w}} = 2 \mathbf{X}^\top_\text{e} \mathbf{X}_\text{e} \mathbf{w} - 2 \mathbf{X}^\top \mathbf{y}  - 2 \mathbf{X}_\text{u} (\mathbf{q} m - \mathbf{q} n + n ) + 2 \lambda \mathbf{w}
$$
and
$$
\frac{\nabla J_r}{\nabla q_i} = m^2 - n^2 - 2 (m-n) \mathbf{x}_i^\top \mathbf{w} \,.
$$
More specifically, in the encoding used here:
$$
\frac{\nabla J_r}{\nabla q_i} = - 2 (\mathbf{x}_i^\top \mathbf{w}-0.5) \,.
$$
If the responsibilities are fixed, this objective function leads to the same solution as the label based objective. Unlike for the label based objective, the objective function is now linear in the responsibilities. Minimizing the responsibilities for a given $\mathbf{w}$ leads to setting $q_i=1$ if  $\mathbf{x}_i^\top \mathbf{w}>0.5$ and $q_i=0$ if $\mathbf{x}_i^\top \mathbf{w}<0.5$

Unlike in the previous approach, the optimal procedure becomes to assign hard labelings to each object when $\mathbf{w}$ is kept fixed.

\section{Optimization}
A straightforward approach to find a local minimum of these functions is to use a block coordinate descend procedure where we iteratively 1) update $\mathbf{w}$, using the closed form solution given while keeping the labels/responsibilities fixed and 2) update $\mathbf{u}$ or $\mathbf{q}$ by the updates derived in the previous section. This leads respectively to a soft-label self-learning and hard-label self-learning procedure.

The convergence of these procedure can be verified by the same argument that was used in \cite{Healy1956}: both steps are guaranteed to decrease the value of the objective function.

While this may seem like a naive approach, block-coordinate descent is the bread and butter of semi-supervised learning approaches. We see this in the EM algorithm, which iteratively updates the responsibilities and the model parameters. But even the seemingly unrelated Transductive SVM \cite{Joachims1999}, in its algorithm iteratively updates imputed labels and the decision function to converge to a local optimum of its objective function.

\section{Difficulty of the Optimization problem}
A nice property of the supervised least squares objective function is that it is convex and allows for a simple closed-form solution. To study the convexity of the semi-supervised extensions of this loss function, we check whether their Hessians are positive semi-definite matrices. The Hessian of the label based extension is given by:
$$
H=\begin{bmatrix} 
2 \mathbf{X}_\text{e}^\top \mathbf{X}_\text{e} & 
-2 \mathbf{X}_\text{u}^\top \\
-2 \mathbf{X}_\text{u} &
-2 \mathbf{I}
\end{bmatrix} \,,
$$
which is not positive semi-definite as it has negative values on its diagonal.
% so for the problem to be convex we would need:
% $$
% 2 \mathbf{z}^\top_1 \mathbf{X}_\text{e}^\top \mathbf{X}_\text{e} \mathbf{z}_1 - 2 \mathbf{z}_2^\top \mathbf{X}_\text{u} \mathbf{z}_1 -2 \mathbf{z}_2^\top \mathbf{I} \mathbf{z}_2  \geq 0 \,.
% $$
% For all $\mathbf{z} \in \mathbb{R}^{d+U}$. Let $\mathbf{z}_1 \in \mathbb{R}^{d}$ be the first $d$ elements in $\mathbf{z}$ and $\mathbf{z}_2 \in \mathbb{R}^{U}$ the remaining $U$ elements. With $\mathbf{z}_1=\mathbf{0}$ and $\mathbf{z}_2 \neq \mathbf{0}$, we have an example where the constraint is violated, showing the function is not convex.

For the responsibility based loss function, the Hessian is slightly different:
% \begin{align}
% \frac{\nabla J}{\nabla \mathbf{q} \nabla \mathbf{q}} = & \mathbf{0} \\
% \frac{\nabla J}{\nabla \mathbf{w} \nabla \mathbf{w}} = & 2 \mathbf{X}_\text{e}^\top \mathbf{X}_\text{e} \\
% \frac{\nabla J}{\nabla \mathbf{w} \nabla \mathbf{u}} = & -2 (m-n) \mathbf{X}_\text{u} \\
% \end{align}
$$
H=\begin{bmatrix} 
2 \mathbf{X}_\text{e}^\top \mathbf{X}_\text{e} & 
-2 (m-n)\mathbf{X}_\text{u}^\top \\
-2 (m-n) \mathbf{X}_\text{u} &
\mathbf{0}
\end{bmatrix} \,.
$$
So for the problem to be convex we would need:
$$
2 \mathbf{z}^\top_1 \mathbf{X}_\text{e}^\top \mathbf{X}_\text{e} \mathbf{z}_1 - 2 (m-n)  \mathbf{z}_2^\top \mathbf{X}_\text{u} \mathbf{z}_1 \geq 0
$$  
for all $\mathbf{z}_1 \in \mathbb{R}^{d}$ and $\mathbf{z}_2 \in \mathbb{R}^{U}$ . When $\mathbf{X}_\text{u}\neq 0$ and $\mathbf{z}_1\neq 0$  it is always possible to pick some $\mathbf{z}_2$ such that the constraint does not hold, so this function is typically not convex either.

Similar to EM-based approaches, we lose convexity for both semi-supervised extensions. One of the extensions may still be easier to optimize than the other. We illustrate the difference between the two objective functions by applying both approaches to the simple dataset shown in Figure~\ref{fig:attractionexample}. We randomly generate $100$ starting values around the supervised solution and plot the convergence of the block coordinate descent procedure for both algorithms in Figure~\ref{fig:attraction1d}. The red line indicates the path of solution when starting at the supervised solution. For the soft-label objective, all starting values converge to the same optimum, while for the hard-label approach, the algorithm converges to four different optima, corresponding to assigning all objects to one of the classes and the two assignments of the clusters to the different classes.

\section{Experiments \& Results}

\subsection{Why soft-labeling can lead to improved predictions}
For hard self-labeling, it is clear why the semi-supervised solution will generally be different from the supervised procedure: for a lot of classifiers it is unlikely that the minimizer of the loss after assigning labels to the unlabeled objects is going to be exactly the same as the minimizer of the loss on the supervised objects. In the soft-label case, it is less clear: because the labeling is more fine-grained, it may be possible to find some assignment of the labels such that the loss does not change. Figure~\ref{fig:simple-example} illustrates when an update of the classifier occurs: starting with the supervised decision function (solid line), we can set the label of the unlabeled object on the left to $0$, such that its contribution to the loss function is $0$. For the object on the right, however, the best we can do is setting the label to $1$. In the next step, the classifier is updated, after which the labels are updated again. This time, the optimal labeling for the right object is again 1, while the optimum for the left object has now changed due to the updated decision function. The effect is that even as the changes are brought about by objects that have decision values greater than $1$ or smaller than $0$, updates lead to a weight vector that has not only changed in magnitude, but the location of the decision boundary has also shifted based on the configuration of the unlabeled objects.

\begin{knitrout}
\definecolor{shadecolor}{rgb}{0.969, 0.969, 0.969}\color{fgcolor}\begin{figure}
\includegraphics[width=\maxwidth]{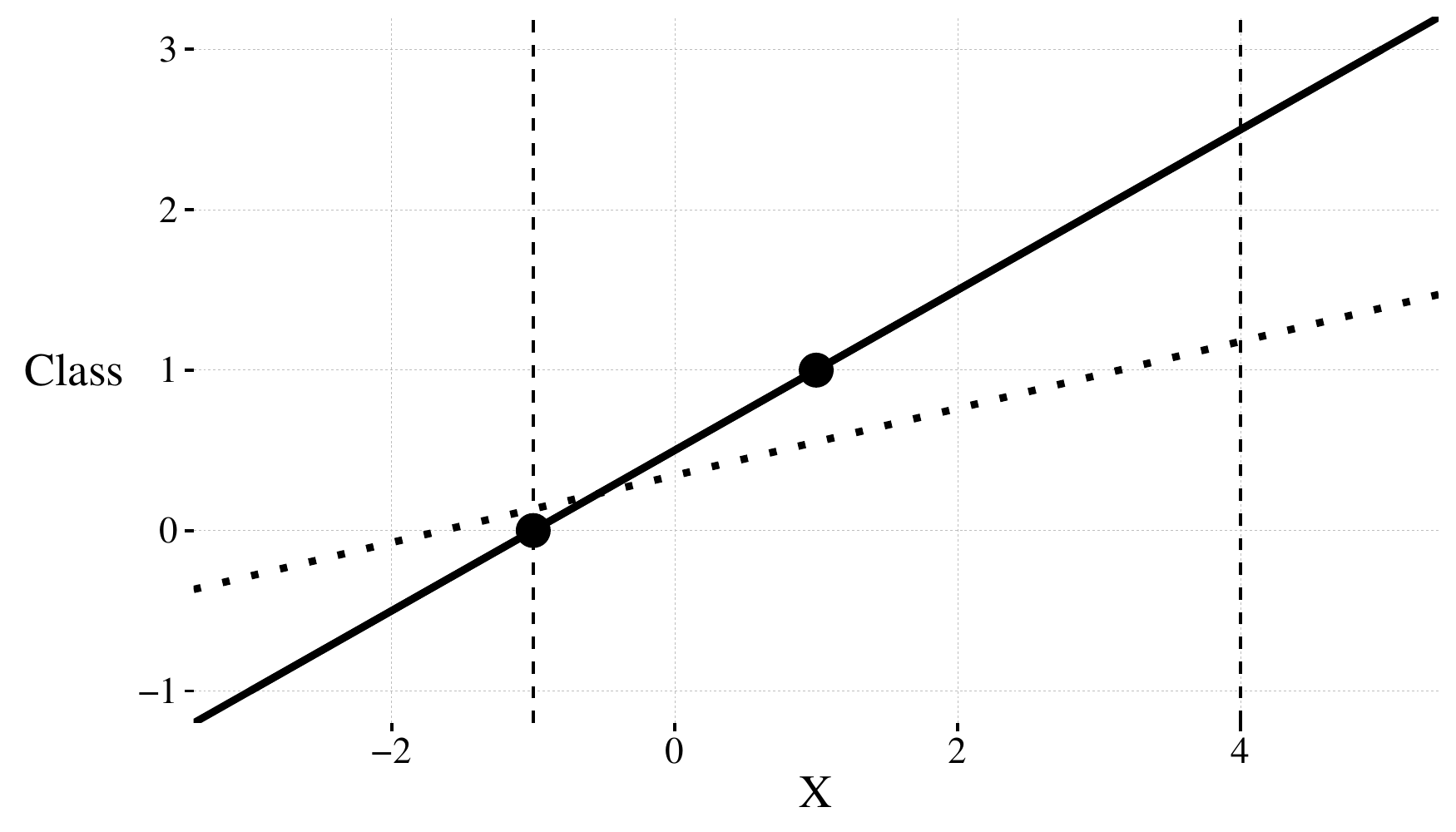} \caption[Example of the first step of soft-label self-learning]{Example of the first step of soft-label self-learning. The circles indicate two labeled objects, while the dashed vertical lines indicate the location of two unlabeled objects. The solid line is the supervised decision function. The dotted line indicates the updated decision function after finding soft labels that minimize the loss of the supervised solution and using these labels as the labels for the unlabeled data in the next iteration.}\label{fig:simple-example}
\end{figure}

\end{knitrout}
 
\begin{knitrout}
\definecolor{shadecolor}{rgb}{0.969, 0.969, 0.969}\color{fgcolor}\begin{figure}
\includegraphics[width=\maxwidth]{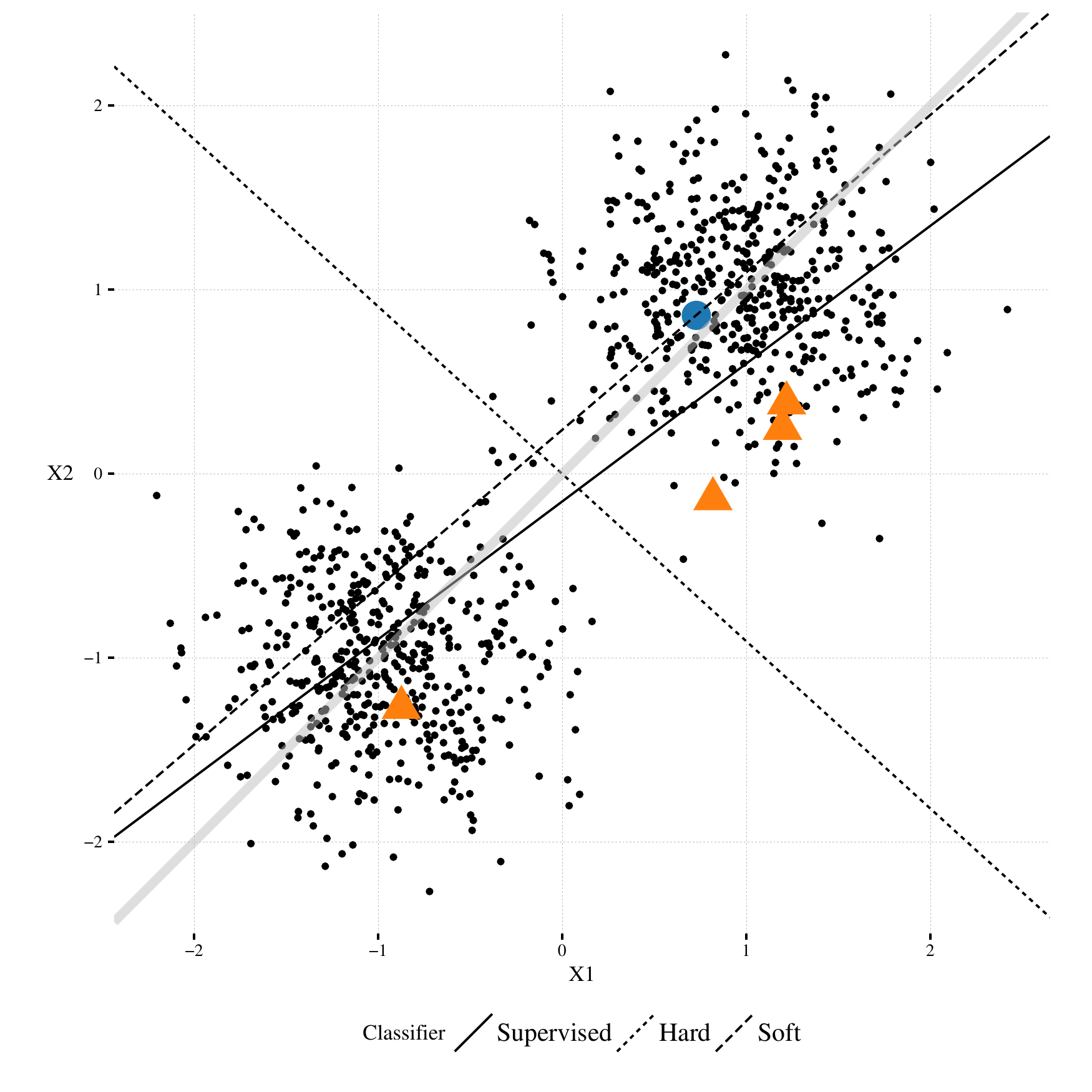} \caption[Example where hard-label self-learning decreases classification performance compared to the supervised solution, while comparable performance to the supervised solution is obtained by soft-label self-learning]{Example where hard-label self-learning decreases classification performance compared to the supervised solution, while comparable performance to the supervised solution is obtained by soft-label self-learning. Light-grey line indicates true decision boundary.}\label{fig:example}
\end{figure}

\end{knitrout}

\subsection{Why hard-labeling might fail}
While Figure~\ref{fig:attraction1d} indicated that the hard-label approach can more easily get stuck in a bad local optimum, using the supervised solution as the initialization leads to a reasonable solution. Figure~\ref{fig:example} shows an example where this is not the case, and the hard-label approach leads away from the reasonable supervised solution to one that has worse performance. The soft-label procedure, on the other hand, gives a classifier that is similar to the supervised procedure.

\subsection{Local optima}
One might wonder how often these bad local minima occur on benchmark datasets. We take a set of $16$ well-known datasets from \cite{Chapelle2006} and the UCI repository \cite{Lichman2013}. Figure~\ref{fig:localoptima} shows the classification performance of the solutions the algorithms converge to for several random initialization of the algorithm and when the algorithms are initialized using the supervised solution. For each dataset, we randomly assign $20\%$ of the objects to the test set, and randomly remove labels from $80\%$ of the remaining objects. We randomly initialize the algorithms $50$ times. In all experiments, $\lambda=0$. The figure shows that the soft-label approach reaches far fewer local optima, compared to the hard-label approach.

\begin{knitrout}
\definecolor{shadecolor}{rgb}{0.969, 0.969, 0.969}\color{fgcolor}\begin{figure}
\includegraphics[width=\maxwidth]{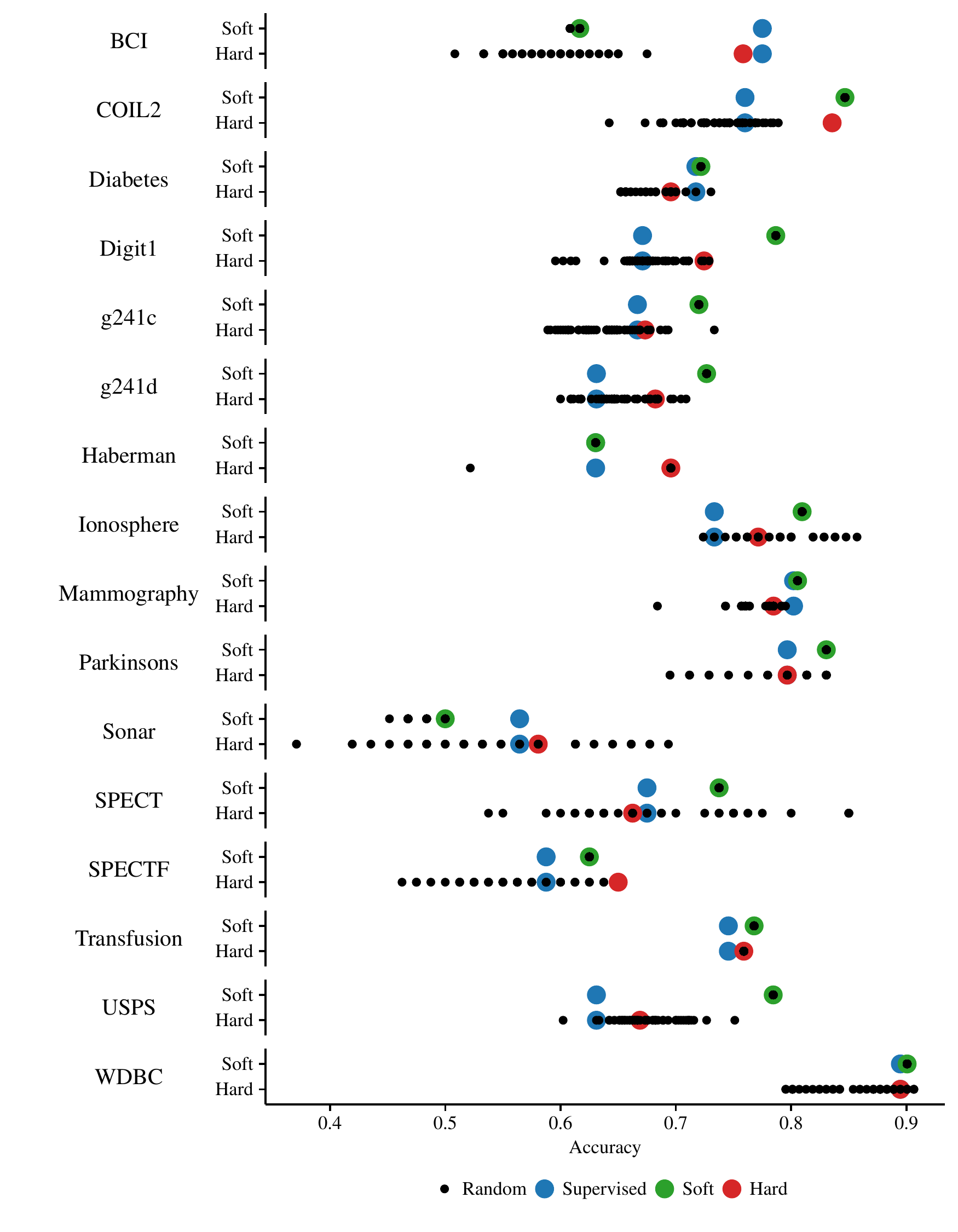} \caption[Each black dot represents the performance of a randomly initialized run of the algorithm]{Each black dot represents the performance of a randomly initialized run of the algorithm. The red points indicate the performance of the hard-label procedure, initialized in the supervised solution. The green points indicate the performance of the soft-label procedure, initialized in the supervised solution. Blue points indicate the performance of the supervised solution. Notice the difference in the number of unique minima of the two procedures.}\label{fig:localoptima}
\end{figure}

\end{knitrout}

\subsection{Increasing the number of unlabeled examples}
We study the effects of using an increasing number of unlabeled examples, while keeping the number of labeled examples fixed. The procedure is set us as follows: for each dataset we sample $L$ objects without replacement to form the labeled set. We choose $L>d$, the number of features in the dataset to ensure the supervised least squares solution is well-defined. We then sample an increasing number of unlabeled examples ($U=1,2,4,\dots,256$) without replacement, and use the remaining samples as the test set. Each classifier is trained on the labeled and unlabeled examples and their classification performance is evaluated on the test set. This procedure is repeated $1000$ times. Apart from the soft- and hard-label self-learning procedures, we also consider an \emph{Oracle} least squares solution, which has access to the labels of the unlabeled objects. This can be considered an upper bound on the performance of any semi-supervised learning procedure. The results are shown in Figure~\ref{fig:learningcurves}.

\begin{knitrout}
\definecolor{shadecolor}{rgb}{0.969, 0.969, 0.969}\color{fgcolor}\begin{figure*}
\includegraphics[width=\maxwidth]{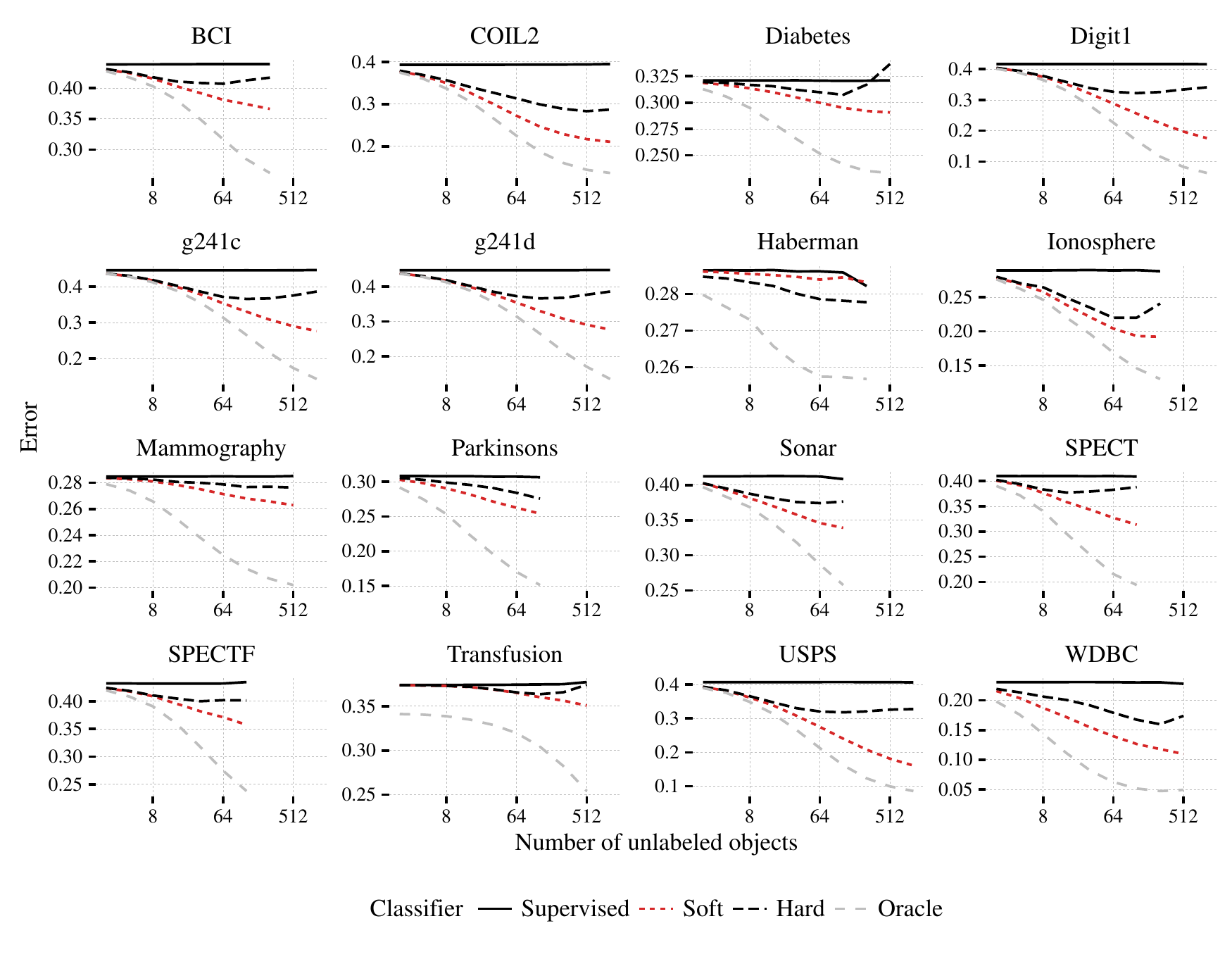} \caption[Classification errors for increasing amounts of unlabeled data]{Classification errors for increasing amounts of unlabeled data. The number of labeled objects remains fixed at a number larger than the dimensionality of the dataset to ensure the supervised solution is well-defined. Results are averaged over $1000$ repeats.}\label{fig:learningcurves}
\end{figure*}

\end{knitrout}
The soft-label self-learning approach generally outperforms the hard-label approach when we increase the amount of unlabeled data. On this collection of datasets, the hard-label variant only outperforms the soft-label variant on the Haberman dataset. Moreover, for the hard-label variant performance can sometimes decrease as more unlabeled data is added. A partial explanation of the better behaviour of the soft-label variant can be found in the results from Figure~\ref{fig:localoptima}. These results suggest that while the hard-label objective has good local minima, there are many more local minima than in the soft-label case, making it less likely to reach a good one when starting from the supervised solution.

\section{Conclusion}
We studied a simple, straightforward semi-supervised least squares classifier that is related to an iterative method that dates back at least 60 years. We described it both in its historical context and by formulating it as a block coordinate descent procedure applied to a particular objective function. The resulting procedure can be considered a soft-label variant of self-learning and is an optimistic type of semi-supervised learning, as a contrast to pessimistic approaches to safe semi-supervised learning. We empirically showed that this simple procedure works well as an alternative to the more fickle hard-label self-learning. Generally, both procedures outperform the supervised solution.

\section*{Acknowledgment}
This work was funded by project P23 of the Dutch public/private research network COMMIT.

\bibliographystyle{IEEEtran}
%\bibliography{IEEEabrv,library}
\bibliography{library}

\end{document}